\documentclass[oneside, 11pt]{article}

\usepackage[utf8]{inputenc} % allow utf-8 input
\usepackage[T1]{fontenc}    % use 8-bit T1 fonts
\usepackage[hidelinks]{hyperref}       % hyperlinks
\usepackage{url}            % simple URL typesetting
\usepackage{booktabs}       % professional-quality tables
\usepackage{amsfonts}       % blackboard math symbols
\usepackage{nicefrac}       % compact symbols for 1/2, etc.
\usepackage{microtype}      % microtypography
\usepackage{amsmath}
\usepackage{amssymb}
\usepackage{graphicx}
\usepackage{bm}
\usepackage{color}
\usepackage{subfigure}
\usepackage{natbib}
\usepackage{multirow}
\usepackage{parskip}

\usepackage[top=1.5in, bottom=1.5in, left=1in, right=1in]{geometry}

\usepackage{lmodern} % REQUIRED TO COMPILE ON WIL PC
\newcommand{\boldSigma}{\bm{\Sigma}}
\newcommand{\boldt}{\mathbf{t}}

\newcommand{\boldy}{\mathbf{y}}

\newcommand{\boldf}{\mathbf{f}}

\newcommand{\boldK}{\mathbf{K}}

\DeclareMathOperator{\cov}{cov} % Operator for the covariance
\DeclareMathOperator{\ex}{\mathbb{E}} % Operator for the expected value
\begin{document}

\title{Non-linear process convolutions for multi-output Gaussian processes}

\author{Mauricio~A~\'{A}lvarez$^1$\thanks{Corresponding author: \texttt{mauricio.alvarez@sheffield.ac.uk}.\newline MAA and WW have been financed by the Engineering and Physical Research Council (EPSRC) Research Project EP/N014162/1. MAA has also been financed by the EPSRC Research Project EP/R034303/1.  CG would like to thank Convocatoria 567 from Administrative Department of Science, Technology and Innovation of Colombia (COLCIENCIAS) for the financial support.}~\quad~Wil~O~C~Ward$^1$~\quad~Cristian~Guarnizo$^2$\\[1em]
$^1$Department of Computer Science, University of Sheffield, UK\\
$^2$Faculty of Engineering, Universidad Tecnol\'ogica de Pereira, Colombia}
\date{{}}

\maketitle

\begin{abstract}
\noindent The paper introduces a non-linear version of the process convolution formalism for building covariance
  functions for multi-output Gaussian processes. The non-linearity is introduced via Volterra series, one
  series per each output. We provide closed-form expressions for the mean function and the covariance
  function  of the approximated Gaussian process at the output of the Volterra series. The mean function and
  covariance function for the joint Gaussian process are derived using formulae for the product moments of
  Gaussian variables. We compare the performance of the non-linear model against the classical process
  convolution approach in one synthetic dataset and two real datasets.
\end{abstract}

\section{INTRODUCTION}

A multi-output Gaussian process (MOGP) is a Gaussian process (GP) with a covariance function that accounts for dependencies between multiple and related outputs \citep{Bonilla:multi07}.
Having models that exploit such dependencies is particularly important when
some of the outputs are expensive to measure and the other more inexpensive outputs can be used as surrogates of the expensive output to
improve its prediction. A typical example comes from geostatistics, where the accuracy of predicting the concentration of toxic heavy metals like lead or
copper, which can be expensive to measure, can be improved by including
measurements of pH as secondary variables, something that is significantly less expensive to measure \citep{Goovaerts:book97}.

One of the challenges in multi-output GPs is defining a cross-covariance function between outputs that leads to a valid covariance function for
the joint GP. There is extensive literature looking at ways to build such types of cross-covariance functions \citep{alvarez2012kernels}. One such approach is known as
process convolution, for which each output is the convolution integral between a smoothing kernel and a latent random process. The approach was introduced
by \citet{Barry:balckbox96}  to build covariance functions for single-output GPs, and later for multi-outputs in \citet{verHoef:convolution98} and \citet{Higdon:convolutions02}.
The convolution integral linearly transforms the underlying latent process, which is usually assumed to be a Gaussian process. The output process is then a GP
with a covariance equal to the convolution operators acting to modify the covariance function of the latent GP.

The main contribution in this paper is the introduction of a non-linear version of the process convolution construction suitable both for single-output and multi-output GPs. The
non-linear model is constructed using a Volterra series where the input function is a latent random process. The Volterra series has been widely studied in the literature of
non-linear dynamical systems \citep{Haber:1999}. They generalise the Taylor expansion for the case of non-instantaneous input functions. We treat the latent process as a Gaussian process and, using formulae for the product moments of Gaussian variables, we provide closed-form expressions for the mean function and covariance function of the output
process. We approximate the output as a Gaussian process using these mean and covariance functions.

Most attempts to generate non-linear models that involve Gaussian processes come from an alternative representation of the convolution integral based on state
space approaches \citep{Hartikainen2012, Sarkka2013}.  Exceptions include the works by \citet{Lawrence2006} and \citet{Titsias2009}  where the non-linearity is a
\emph{static transformation} of the underlying latent GP. We review these and other approaches in the Section~\ref{sec:relatedwork}.

We compare the performance of the non-linear model against the classical process convolution approach in one synthetic dataset and two real datasets and show that the
non-linear version provides better performance than the traditional approach.

\section{BACKGROUND}

In this section, we briefly review the MOGP with process convolutions. We will refer to this particular construction as the convolved MOGP (CMOGP). We also briefly
review the Volterra series and the formulae for the product moments of Gaussian variables that we use to construct our non-linear model.

\subsection{Process convolutions for multi-output Gaussian processes}

We want to jointly model the set of processes $[f_d(t)]_{d=1}^D$ using a joint Gaussian process. In the process convolution approach to building such a
Gaussian process, dependencies between different ouputs are introduced by assuming that each output $f_d(t)$ is the convolution integral between
a smoothing kernel $G_d(t)$ and some latent function $u(t)$,
\begin{align}\label{eq:conv:integral}
f_d(t)= \int_\tau G_d(t-\tau)u(\tau)d\tau,
\end{align}
where the smoothing kernel $G_d(t-\tau)$ should have finite energy. Assuming that the latent process $u(t)$ is a GP with zero mean function and covariance
function $k(t,t')$, the set of processes $f_1(t), f_2(t), \cdots, f_{D}(t)$ are jointly Gaussian with zero mean function and cross-covariance function between $f_d(t)$
and $f_{d'}(t')$ given by
\begin{align}k_{d,d'}(t,t') = \cov[ f_{d}(t), f_{d'}(t')] = \iint G_d(t-\tau_i) G_{d'}(t'-\tau_i) k(\tau_i,\tau_j)\text{d}\tau_{i}\text{d}\tau_{j}.\label{eq:crosscov_linear_case}
\end{align}

In \citet{alvarez2012kernels}, the authors have shown that the covariance function above generalises the well-known linear model of coregionalization,
a form of covariance function for multiple outputs commonly used in machine learning and geostatistics.

The expression for $f_d(t)$ in the form of the convolution integral in  \eqref{eq:conv:integral} is also the representation of a linear dynamical system with
impulse response $G_d(t)$. In the context of latent force models, such convolution expressions have been used to compute covariance functions informed by
physical systems where the smoothing kernel is related to the so-called Green's function representation of an ordinary differential operator
\citep{Alvarez:lfm:pami:2013}.

\subsection{Representing non-linear dynamics with Volterra series}
We borrow ideas from the representation of non-linear systems to extend the CMOGP to the non-linear case. One such representation is the Taylor series, which is the expansion of a time-invariant non-linear system as a polynomial about a fixed working point:
\begin{equation*}
    f(t) = \sum^\infty_{c=0}g_iu^c(t) = g_0 + g_1u(t) + g_2u^2(t) + \ldots
\end{equation*}
While the Taylor series is widely used in the approximation of non-linear systems, it can only approximate systems for which the input has an instantaneous
effect over the output \citep{Haber:1999}.

By the Stone-Weierstra\ss{} theorem, a given continuous non-linear system with finite-dimensional vector input can be uniformly appoximated by a finite polynomial series \citep{Gallman1976}. The Volterra series is such a polynomial expansion, describing a series of nested convolution operators:
\begin{align*}
\begin{split}
f(t) &= \sum^\infty_{c=0} \int\ldots\int G^{(c)}(t-\tau_1,\ldots,t-\tau_u)\prod^c_{j=1}u(\tau_j)\mathrm{d}\tau_j\\
&= G^{(0)} + \int G^{(1)}(t-\tau_1)u(\tau_1)\mathrm{d}\tau_1 + \iint G^{(2)}(t-\tau_1,t-\tau_2)u(\tau_1)u(\tau_2)\mathrm{d}\tau_1\mathrm{d}\tau_2 + \ldots
\end{split}
\end{align*}
The leading term $G^{(0)}$ is a constant term, which in practise is assumed to be zero-valued and the series is
incremented from $c=1$. Because of the convolutions involved, the series is no longer modelling an instantaneous input at $t$, giving the series a so-called \emph{memory effect} \citep{Haber:1999}. As with the Taylor series, the approximant needs a cut-off for the infinite sum, denoted $C$; a Volterra series with $C$ sum terms is called $C$-order.

The representation of a $c^\mathrm{th}$ degree kernel, i.e. $G^{(c)}(t-\tau_1,\ldots,t-\tau_c)$, can be expressed in different forms, such as in symmetric or triangular form \citep{Haber:1999}. A common assumption is that the kernels are homogeneous and seperable, such that $G^{(c)}$ is the product of $c$ first degree kernels. The assumption of separability is stronger but reduces the number of unique parameters, which can be very large for a full Volterra series \citep{Schetzen1980}. It should also be noted that a truncated Volterra series with seperable homogeneous kernels is equivalent to a Wiener model \citep{Cheng2017}.

\subsection{Product moments for multivariate Gaussian random variables}\label{sec:prod:moments}

Several of the results that we will present in the following section involve the computation of the expected value of the product of several multivariate Gaussian random variables. For this, we will use results derived in \citet{Song:ProductMoments:2015}. We are interested in those results for which the Gaussian random variables have zero mean. In particular, let $[X_i]_{i=1}^c$ be multivariate Gaussian random variables with zero mean values and covariance between $X_i$ and $X_j$ given as $\phi_{ij}$. According to Corollary 1 in \citet{Song:ProductMoments:2015}, the expression for $\ex\left[\prod_{i=1}^{c}X_i^{a_i}\right]$, follows as
\begin{align}\label{prod:moment:gaussians:a:gt:one}
    \ex\bigg[\prod_{i=1}^c X^{a_i}_i\bigg] & = \sum_{\mathbf{L}\in T_{\mathbf{a}}}\frac{\prod_{k=1}^ca_k!}{2^{\mathrm{tr}[{\mathbf{L}}]}\prod_{i=1}^c\prod_{j=1}^cl_{ij}!}\prod_{i=1}^c\prod_{j=i}^c(\phi_{ij})^{l_{ij}},
\end{align}
%
%#########
% ORIGINAL EXPLANATION
% \/
% where $\mathbf{l} = \{\{l_{ij}\}_{i=1}^c\}_{j=i}^c$, $\mathbf{a} =\{a_k\}_{k=1}^c$ and the elements in $\mathbf{l}$ and $\mathbf{a}$, $l_{ij}$ and $a_k$, are nonnegative integers; $T_{\mathbf{a}}$ denotes the collection of all vectors $\mathbf{l}=\{l_{ij}\}_{i=1,j=1}^c$ ($l_{ij}=l_{ji}$) such that $\{L_{\mathbf{a},k}=0\}_{k=1}^c$, with $L_{\mathbf{a},k}= a_k-l_{kk}-\sum_{j=1}^cl_{jk}$; and $M_{\mathbf{l}} = \sum_{k=1}^cl_{kk}$.
%
%##########
% REVISED NOTATION
% \/
where $\mathbf{a} = [a_i]^c_{i=1}$ is a vector consisting of the random variable exponents and $T_{\mathbf{a}}$ is the set of $c\times c$ \emph{symmetric} matrices\footnote{In \citet{Song:ProductMoments:2015}, the authors denote $T_{\mathbf{a}}$ as a collection of sets-of-sets, but we interpret the elements as symmetric matrices for notational clarity.} that meet the condition $L_{\mathbf{a},k} = 0$ for $k=1,\ldots,c$, as defined by
\small{\begin{align}
    T_\mathbf{a} = \bigg\{[l_{pq}]_{c\times c}\,\bigg|\,\underbrace{a_k - l_{kk} - \sum^c_{j=1}l_{jk}}_{L_{\mathbf{a},k}} = 0,\,\forall k\in\{1,\ldots,c\}\bigg\}.\label{set:conditional:productmoment}
\end{align}}%
%
%====
If the sum of exponents, $\sum_{k=1}^ca_k$, is an odd number, then  $\ex[\prod_{i=1}^{c}X_i^{a_i}]=0$ for the zero mean value case, as described in Corollary 2 in \citet{Song:ProductMoments:2015}.

An additional result used later is that if $a_k = 1$, $\forall a_k$ then \eqref{prod:moment:gaussians:a:gt:one}, by Remark 5 in \citet{Song:ProductMoments:2015}, reduces to
\begin{align}\label{prod:moment:gaussians:a:equal:one}
  \ex\left[\prod_{i=1}^{c}X_i \right] = \sum_{\mathbf{L}\in T_{\mathbf{a}}}\prod_{i=1}^{c}\prod_{j=i}^{c}\phi_{ij}^{l_{ij}}.
\end{align}

\section{A NON-LINEAR CMOGP BASED ON VOLTERRA SERIES}

We represent a vector-valued non-linear dynamic system with a system of Volterra series of order $C$. For a given output dimension, $d$, we approximate the function with a truncated Volterra series as follows
{\small%
\begin{align}
    f_d^{(C)}(t) = \sum_{c=1}^C \int\cdots\int G^{(c)}_d(t-\tau_1, \ldots, t-\tau_c) \prod_{j=1}^c u(\tau_j) \text{d}\tau_j,\label{eq:finite:ncmogp}
\end{align}}%
where $G^{(c)}_d$ are $c^\text{th}$ degree Volterra kernels.

Our approach is to assume that $u(t)$, the latent driving function, follows a GP prior. For $C=1$, we recover the the expression for the process convolution construction of a multi-output GP as defined in \eqref{eq:conv:integral}. In contrast to the linear case, the output $f_d^{(C)}$ is no longer a GP. However, we approximate $f^{(C)}_d$ with a GP $\tilde{f}^{(C)}_d(t)$ with a mean and covariance function computed from the moments of the output process $f^{(C)}_d$:
\begin{align}
    \tilde{f}^{(C)}_d(t) \sim \mathcal{GP}(\mu^{(C)}_d(t), k^{(C)}_{d,d'}(t,t')),
\end{align}
where $\mu^{(C)}_d(t) = \ex[f^{(C)}_d(t)]$ and $k^{(C)}_{f_df_{d'}}(t,t') = \cov[f^{(C)}_d(t),f^{(C)}_{d'}(t')]$. Approximating a non-Gaussian distribution with a Gaussian, particularly for non-linear systems, is common in state space modelling, for example in the unscented Kalman filter \citep{Sarkka2013b}; or as a choice of variational distribution in variational inference \citep{Blei2017}. We refer to the joint process $[\tilde{f}^{(C)}_d(t)]^D_{d=1}$ as the \emph{non-linear convolved multi-output GP} (NCMOGP).
%==

Furthermore, we will assume that the $c^\text{th}$ degree Volterra kernels are separable, such that
\begin{align}
  G^{(c)}_d(t-\tau_1, \ldots, t-\tau_c) = \prod_{i=1}^c G^{(c,i)}_d(t-\tau_i),\label{eq:separable_volterra_kern}
\end{align}
where $G^{(c,i)}_d$ are first degree Volterra kernels.

Using this separable form, we express the output $f^{(C)}_d(t)$ as
\begin{align*}
\sum_{c=1}^{C} \prod_{i=1}^c \int_{\tau_i}G^{(c,i)}_d(t-\tau_i) u(\tau_{i})\text{d}\tau_{i}= \sum_{c=1}^{C} \prod_{i=1}^c f^{(c,i)}_d(t),
\end{align*}
where we define
\begin{align*}
  f^{(c,i)}_d(t) & =\int_{\tau_i}G^{(c,i)}_d(t-\tau_i) u(\tau_{i})\text{d}\tau_{i}.
\end{align*}
Assuming $u(t)$ has a GP prior with zero mean and covariance $k(t,t')$, and due to the linearity of the expression above, we can compute the corresponding mean and covariance functions for the joint Gaussian process $[f^{(c,i)}_d(t)]_{d=1}^D$. We compute the cross-covariance function between $ f^{(c,i)}_d(t)$ and $ f^{(c',j)}_{d'}(t')$ using
\begin{align}
    &k^{(c,i),(c',j)}_{d,d'}(t,t') = \cov[ f^{(c,i)}_{d}(t), f^{(c,j)}_{d'}(t')] = \iint G^{(c,i)}_d(t-\tau_i) G^{(c',j)}_{d'}(t'-\tau_i) k(\tau_i,\tau_j)\text{d}\tau_{i}\text{d}\tau_{j}.\label{eq:crosscov_kof_ffprime}
\end{align}
This is a similar expression to the one in \eqref{eq:crosscov_linear_case} for the CMOGP. For some particular forms of the Volterra kernels $G^{(c,i)}_d$ and covariance $k(t,t')$ of the latent process $u(t)$, the covariance $k^{(c,i),(c',j)}_{d, d'}(t,t')$ can be computed analytically.
%%##==
\subsection{NCMOGP with separable Volterra kernels}\label{sec:sep}
%#=
In this section, we derive expressions for $\mu^{(C)}_d(t)$ and $k^{(C)}_{d,d'}(t,t')$ with the assumption of separability of the Volterra kernels in \eqref{eq:separable_volterra_kern}.

\subsubsection{Mean function}\label{sec:mean:sep}
Let us first compute the mean function, $\mu^{(C)}_d(t) = \ex[f^{(C)}_d(t)]$. Using the definition for the expected value, we get
\begin{align}
 \ex[ f^{(C)}_d(t)] & =\sum_{c=1}^{C} \ex\Bigg[\prod_{i=1}^c f^{(c,i)}_d(t)\Bigg].\label{mean:ncmogp:separable}
\end{align}

The expected value of the product of the Gaussian processes, $\ex[\prod_{i=1}^c f^{(c,i)}_d(t)]$, can be computed using results obtained for the expected value of the product of multivariate Gaussian random variables as introduced in Section~\ref{sec:prod:moments}.

Applying the result in \eqref{prod:moment:gaussians:a:equal:one} to the expression of the expected value in \eqref{mean:ncmogp:separable}, we get
\begin{align}\begin{split}
\ex\Bigg[\prod_{i=1}^c f^{(c,i)}_d(t)\Bigg] & = \sum_{\mathbf{L}\in T_{\mathbf{a}}}\prod_{i=1}^{c}\prod_{j=i}^{c}\bigg( k^{(c,i), (c,j)}_{d, d}(t, t)\bigg)^{l_{ij}},
\end{split}\label{exp:product:seperable}\end{align}
where
\begin{align*}
k^{(c,i), (c,j)}_{d, d}(t, t) = \cov[ f^{(c,i)}_{d}(t), f^{(c,j)}_{d}(t)].
 \end{align*}
Note that only the terms for which $c$ is even are non-zero.

%=?
\paragraph{Example 1} To see an example of the kind of expressions that the expected value takes, let us assume that $c=4$. We then have
\begin{align*}
\ex\Bigg[\prod_{i=1}^4 f^{(4,i)}_d(t)\Bigg] & = \sum_{\mathbf{L}\in T_{\mathbf{a}}}\prod_{i=1}^{4}\prod_{j=i}^{4}\bigg( k^{(4,i), (4,j)}_{d, d}(t, t)\bigg)^{l_{ij}},
 \end{align*}
where
\begin{align*}
k^{(4,i), (4,j)}_{d, d}(t, t) = \cov[ f^{(4,i)}_{d}(t), f^{(4,j)}_{d}(t)].
 \end{align*}
We now need to find the set $T_{\mathbf{a}}$ containing all $c\times c$ symmetric matrices $\mathbf{L}$, the elements of which meet the conditon described in \eqref{set:conditional:productmoment}, where $\mathbf{a} = [1\;1\;1\;1]$. This leads to the following system of equations
\begin{align*}
    2l_{11} +  l_{12} + l_{13}+l_{14} &= 1\\
    l_{12} +  2l_{22} + l_{23}+l_{24} &= 1\\
    l_{13} +  l_{23} + 2l_{33}+l_{34} &= 1\\
    l_{14} +  l_{24} + l_{34}+2l_{44} &= 1,
\end{align*}
%======
% ORIGINAL DETAILS
% \/
% where we have used $l_{ij}=l_{ji}$. It can be seen from the set of equations above that the vectors $\mathbf{l}=\{\{l_{ij}\}_{i=1}^4\}_{j=i}^4$ that belong to $T_{\mathbf{a}}$ include a vector $\mathbf{l}_1$ for which only $l_{12}=1, l_{34}=1$ and all the other $l_{ij}=0$; a vector $\mathbf{l}_{2}$ for which only $l_{13}=1, l_{24}=1$ and all the other $l_{ij}=0$; and a vector $\mathbf{l}_{3}$ for which only $l_{14}=1, l_{23}=1$ and all the other $l_{ij}=0$. We can now get an expression for the expected value as follows
%========
% REVISED DESCRIPTON
% \/
where we have used the symmetry of $\mathbf{L}$, so $l_{ij} = l_{ji}$. It can be seen from the above system that the set $T_{\mathbf{a}}$ contains three unique symmetric matrices:
{\small
\begin{align*}
    T_{\mathbf{a}} = \left\{\begin{bmatrix}0 & 1 & 0 & 0 \\ 1 & 0 & 0 & 0 \\ 0 & 0 & 0 & 1 \\ 0 & 0 & 1 & 0\end{bmatrix}, \begin{bmatrix}0 & 0 & 1 & 0 \\ 0 & 0 & 0 & 1 \\ 1 & 0 & 0 & 0 \\ 0 & 1 & 0 & 0\end{bmatrix}, \begin{bmatrix}0 & 0 & 0 & 1 \\ 0 & 0 & 1 & 0 \\ 0 & 1 & 0 & 0 \\ 1 & 0 & 0 & 0\end{bmatrix}\right\}.
\end{align*}}%
We now have an expression for the expected value, by \eqref{exp:product:seperable}:
\begin{align*}
  \ex\Bigg[\prod_{i=1}^4 f^{(4,i)}_d(t)\Bigg] = k^{(4,1), (4,2)}_{d, d}(t, t)  k^{(4,3), (4,4)}_{d, d}(t, t) +k^{(4,1), (4,3)}_{d, d}(t, t)  k^{(4,2), (4,4)}_{d, d}(t, t) +k^{(4,1), (4,4)}_{d, d}(t, t)  k^{(4,2), (4,3)}_{d, d}(t, t).
\end{align*}

%#=
\subsubsection{Cross-covariance function}
For computing the covariance function, $k^{(C)}_{d,d'}(t,t') = \ex[f^{(C)}_d(t)f^{(C)}_{d'}(t')] - \ex[f^{(C)}_d(t)]\ex[f^{(C)}_{d'}(t')]$, we first need to compute the second moment between $f^{(C)}_d(t)$ and $f^{(C)}_{d'}(t')$. The second moment is given as
\begin{align}\begin{split}
    \ex\bigg[f^{(C)}_d(t)f^{(C)}_{d'}(t')\bigg] &=\ex\bigg[\sum_{c=1}^{C} \prod_{i=1}^c f^{(c,i)}_d(t) \sum_{c'=1}^{C} \prod_{j=1}^{c'}f^{(c',j)}_{d'}(t')\bigg]\\
    &= \sum_{c=1}^{C} \sum_{c'=1}^{C} \ex\bigg[\prod_{i=1}^c \prod_{j=1}^{c'} f^{(c,i)}_d(t) f^{(c',j)}_{d'}(t')\bigg] \\
    &= \sum_{c=1}^{C} \sum_{c'=1}^{C} \ex\left[\prod^{c+c'}_{i=1}\bar{f}_{d,d'}^{(i)}(t)\right],
\end{split}\label{second:moment:ncmogp:seperable}\end{align}
where $\bar{f}_{d,d'}^{(i)}$ is the $i^\text{th}$ output of a vector-valued function consisting of all functions in the product
{\small\begin{align*}
\bm{\bar{f}}_{d,d'}(t) = \left[%
    f^{(c,1)}_d(t)\;\ldots\;f^{(c,c)}_d(t)\; f^{(c',1)}_{d'}(t')\;\ldots\; f^{(c',c')}_{d'}(t')\right]^\top.
\end{align*}}%
We have assumed that both $f^{(C)}_d(t)$ and $f^{(C)}_{d'}(t')$ share the same value of $C$, although a more general expression can be obtained where each output can have its own $C$ value. We can apply the expressions in \citet{Song:ProductMoments:2015} to the above moment of the product of Gaussian random variables as we did for computing the mean function in Section~\ref{sec:mean:sep}. Using the expression for the expected value of the product of Gaussian variables, we get
%
%======
% ORIGINAL NOTATION
% % \/
% \begin{align*}
%   \ex\bigg[&\prod_{i=1}^c \prod_{j=1}^{c'} f^{(c,i)}_d(t) f^{(c',j)}_{d'}(t')\bigg]\\
%                  &=   \sum_{\mathbf{l}\in T_{\mathbf{a}}} \prod_{i=1}^{c+c'}
%   \prod_{j=i}^{c+c'}\bigg( k^{(\permc(i),\permp(i)), (\permc(j),\permp(j))}_{\permo(i), \permo(j)}(\permi(i), \permi(j))\bigg)^{l_{ij}},
% \end{align*}
% where
% \begin{align*}
%   &k^{(\permc(i),\permp(i)), (\permc(j),\permp(j))}_{\permo(i), \permo(j)}(\permi(i), \permi(j))  = \\
%   &\cov[ f^{(\permc(i),\permp(i))}_{\permo(i)}(\permi(i)), f^{(\permc(j),\permp(j))}_{\permo(j)}(\permi(j))],
% \end{align*}
% with
% \begin{align*}
%   \permc & = [\underbrace{c\;\;\cdots\;\;c}_{c\text{ times }}\;\;\underbrace{c'\;\;\cdots\;\;c'}_{c'\text{ times }}],\\
%   \permp & = [1\;\;\cdots\;\;c\;\;1\;\;\cdots\;\;c'],\\
%   \permi  & = [\underbrace{t\;\;\cdots\;\;t}_{c\text{ times }}\;\; \underbrace{t'\;\;\cdots\;\;t'}_{c'\text{ times }}],\\
%   \permo & = [\underbrace{d\;\;\cdots\;\;d}_{c\text{ times }}\;\; \underbrace{d'\;\;\cdots\;\;d'}_{c'\text{ times }}].
% \end{align*}
% The expression for the second moment only includes the terms for which $c+c'$ is even.
%
%========
% NEW NOTATION
% \/
\small{%
\begin{align*}
    \ex\bigg[\prod^{c+c'}_{i=1}\bar{f}_{d,d'}^{(i)}(t)\bigg] = \sum_{\mathbf{L}\in T_{\mathbf{a}}} \prod_{i=1}^{c+c'}
    \prod_{j=i}^{c+c'}\left(\cov\!\left[\bar{f}_{d,d'}^{(i)}(t), \bar{f}_{d,d'}^{(j)}(t)\right]\right)^{l_{ij}},
\end{align*}}%
where the covariance element is defined in \eqref{eq:crosscov_kof_ffprime} as the cross-covariance of two latent functions.
%
%#?
%========
% OLD EXAMPLE
% \/
%
% \paragraph{Example 2} For illustration purposes, let as assume that $c=3$ and $c'=1$. In this case, we would have
% \begin{align*}
%  \ex\bigg[&\prod_{i=1}^3 f^{(c,i)}_d(t) f^{(1,1)}_{d'}(t')\bigg]\\
%                  &=   \sum_{\mathbf{l}\in T_{\mathbf{a}}} \prod_{i=1}^{4}
%   \prod_{j=i}^{4}\bigg( k^{(\permc(i),\permp(i)), (\permc(j),\permp(j))}_{\permo(i), \permo(j)}(\permi(i), \permi(j))\bigg)^{l_{ij}},
% \end{align*}
% with $\permc = [3\; 3\; 3 \;1]$, $\permp = [1\; 2\; 3 \;1]$, $\permi=[t \; t \;t\; t']$, and $\permo = [d \;d \;d\; d']$. The vectors $\mathbf{l}$ in the set $T_{\mathbf{a}}$ are exactly the same ones that we found in Example 1 in Section~\ref{sec:mean:sep} leading to
%========
% REVISED NOTATION EXAMPLE
% \/
\paragraph{Example 2} For illustration purposes, let us assume that $c=3$ and $c' = 1$. In this case, we would have
\begin{align*}
    \ex\bigg[&\prod_{i=1}^3 f^{(3,i)}_d(t) f^{(1,1)}_{d'}(t')\bigg]= \sum_{\mathbf{L}\in T_{\mathbf{a}}} \prod_{i=1}^{4}
    \prod_{j=i}^{4}\left(\cov\!\left[\bar{f}_{d,d'}^{(i)}(t), \bar{f}_{d,d'}^{(j)}(t)\right]\right)^{l_{ij}},
\end{align*}
where we have defined $\bm{\bar{f}}_{d,d'}$ as
\begin{align*}
    \bm{\bar{f}}_{d,d'} = \begin{bmatrix}
        f^{(3,1)}_d(t) & f^{(3,2)}_d(t) & f^{(3,3)}_d(t) & f^{(1,1)}_{d'}(t')
    \end{bmatrix}^\top.
\end{align*}
The set $T_{\mathbf{a}}$ contains the same matrices as we found in Example 1 in Section~\ref{sec:mean:sep} leading to
\begin{align*}
  \ex\bigg[\prod_{i=1}^3 f^{(c,i)}_d(t) f^{(1,1)}_{d'}(t')\bigg]  =&  k^{(3,1), (3,2)}_{d, d}(t, t)  k^{(3,3), (1,1)}_{d, d'}(t, t') \\ &\quad +  k^{(3,1), (3,3)}_{d, d}(t, t)  k^{(3,2), (1,1)}_{d, d'}(t, t') +  k^{(3,1), (1,1)}_{d, d'}(t, t')  k^{(3,2), (3,3)}_{d, d}(t, t).
\end{align*}
The cross-covariance function between $f^{(C)}_d(t)$ and $f^{(C)}_{d'}(t')$ is then computed using
\begin{align*}
  \cov[&f^{(C)}_d(t), f^{(C)}_{d'}(t')] \\
  &=\ex[f^{(C)}_d(t)f^{(C)}_{d'}(t')] - \ex[f^{(C)}_d(t)]\ex[f^{(C)}_{d'}(t')].
\end{align*}
We have expressions for both $\ex[f^{(C)}_d(t)f^{(C)}_{d'}(t')]$ and $\ex[f^{(C)}_d(t)]$ by \eqref{mean:ncmogp:separable} and \eqref{second:moment:ncmogp:seperable} respectively.%, so we can compute this covariance.

\subsection{NCMOGP with separable and homogeneous Volterra kernels}

In the section above, we introduced a model that allows for different first-order Volterra kernels $G^{(c,i)}_d(x)$, when building the general Volterra kernel of order $c$. A further assumption we can make is that all first-order Volterra kernels are the same, i.e.,
$G^{(c,i)}_d(t-\tau_i) = G_d(t-\tau),\quad \forall c, \;\forall i,$ meaning that $f^{(c,i)}_d(t) = f_d(t)$, for all $c$ and for all $i$. We will refer to this as the separable and homogeneous version of the NCMOGP.

We then get \begin{align*}
  f^{(C)}_d(t) & =\sum_{c=1}^{C} \prod_{i=1}^c f^{(c,i)}_d(t) = \sum_{c=1}^{C} \prod_{i=1}^c f_d(t) = \sum_{c=1}^{C} f^{c}_d(t),
\end{align*}
where $ f_d(t) =\int G_d(t-\tau) u(\tau)\text{d}\tau$, and $u(t) \sim \mathcal{GP}(0, k(t,t'))$. The cross-covariance function between $f_d(t)$ and $f_{d'}(t')$ is again $k_{d,d'}(t,t')$ as in \eqref{eq:crosscov_linear_case}.

As we did in Section~\ref{sec:sep}, we can compute the mean function for $f^{(C)}_d(t)$ and cross-covariance functions between $f^{(C)}_d(t)$ and $f^{(C)}_{d'}(t')$.

\subsubsection{Mean function}

Using Eq. \eqref{prod:moment:gaussians:a:gt:one}, the mean function $\ex[f^{(C)}_d(t)]$ follows as
\begin{align*}
  \ex[f^{(C)}_d(t)] &= \ex\bigg[\sum_{c=1}^{C} f^{c}_d(t) \bigg] = \sum_{c=1}^{C} \ex\big[ f^{c}_d(t) \big] = \sum_{c=1}^{C} \frac{c!}{2^{l_{11}}l_{11}!}(k_{d,d}(t,t))^{l_{11}},
\end{align*}
where $l_{11}$ is such that satisfies $2l_{11} = c$. This means that the above expression only has solutions for even-valued $c$, and $l_{11}= c/2$. Then,
\begin{align*}
  \ex[f^{(C)}_d(t)] & = \sum_{c=1}^{C} \frac{c!}{2^{c/2}\left(\frac{c}{2}\right)!}(k_{d,d}(t,t))^{c/2},
\end{align*}
for even $c$ and $C\ge 2$.

\subsubsection{Cross-covariance function}

We can compute the second moment $\ex[f^{(C)}_d(t) f^{(C)}_{d'}(t')]$ using
\begin{align*}
\ex[f^{(C)}_d(t) f^{(C)}_{d'}(t')] = \ex\bigg[\sum_{c=1}^{C} f^{c}_d(t) \sum_{c'=1}^{C} f^{c'}_{d'}(t')\bigg] = \sum_{c=1}^{C} \sum_{c'=1}^{C} \ex\big[ f^{c}_d(t) f^{c'}_{d'}(t')\big].
\end{align*}
Once again we use expression \eqref{prod:moment:gaussians:a:gt:one} to compute $\ex\big[ f^{c}_d(t) f^{c'}_{d'}(t')\big]$, leading to
\begin{align*}
\ex\big[ f^{c}_d(t) f^{c'}_{d'}(t')\big] =  \sum_{\mathbf{L}\in T_{\mathbf{a}}}  A_{c, c',\mathbf{L}}  (k_{d,d}(t,t))^{l_{11}}(k_{d,d'}(t,t'))^{l_{12}}(k_{d',d'}(t',t'))^{l_{22}},
\end{align*}
where $A_{c, c',\mathbf{L}}$ is defined as
\begin{align*}
 A_{c, c',\mathbf{L}} =\frac{c!c'!}{2^{l_{11}+l_{22}}l_{11}!l_{12}!l_{22}!},
\end{align*}
and $k_{d,d'}(t,t')$ is defined in \eqref{eq:crosscov_linear_case}. To
avoid computer overflow due to the factorial operators when computing
$A_{c, c',\mathbf{L}}$, we compute $\exp(\log(A_{c, c',\mathbf{L}}))$
instead. As stated previously, the expected value will be 0 if $c+c'$
is not even.

\paragraph{Example 3} Let as assume that $c=3$ and $c'=3$. The expected value $\ex\big[ f^{3}_d(t) f^{3}_{d'}(t')\big]$ follows as
\begin{align*}
\sum_{\mathbf{L}\in T_{\mathbf{a}}}  A_{3, 3,\mathbf{L}}  (k_{d,d}(t,t))^{l_{11}}(k_{d,d'}(t,t'))^{l_{12}}(k_{d',d'}(t',t'))^{l_{22}},
\end{align*}
where $A_{3, 3,\mathbf{L}}
=(3!3!)/{2^{l_{11}+l_{22}}l_{11}!l_{12}!l_{22}!}$. To find the elements in $\mathbf{L}\in T_{\mathbf{a}}$, we need to solve similar equations
to the ones in Example 1. We would have
\begin{align*}
    2l_{11} +  l_{12}  &= c = 3\\
    l_{12} +  2l_{22}  &= c' =3.
\end{align*}
We can see that there are two valid solutions for $\mathbf{L}$:
\begin{align*}
    T_\mathbf{a} = \left\{\begin{bmatrix}0 & 3 \\ 3 & 0\end{bmatrix},\begin{bmatrix}1 & 1 \\ 1 & 1\end{bmatrix}\right\}.
\end{align*}
The expression for $\ex\big[ f^{3}_d(t) f^{3}_{d'}(t')\big]$ is thus
 \begin{align*}
6 k ^{3}_{d,d'}(t,t') +  9 k_{d,d}(t,t)k_{d,d'}(t,t')k_{d',d'}(t',t').
\end{align*}
We compute the covariance $\cov[f_{d}^{(C)}(t), f_{d'}^{(C)}(t')]$ now that we have expressions to compute
the mean for $f_{d}^{(C)}(t)$ and the expected value for the product between $f_{d}^{(C)}(t)$ and  $f_{d'}^{(C)}(t')$.

\section{RELATED WORK}\label{sec:relatedwork}
In the work by \citet{Lawrence2006}, non-linear dynamics are
introduced with a GP prior within a non-linear function, which are
inferred using the Laplace approximation, with the convolution
operator itself approximated as a discrete sum. Similarly,
\citet{Titsias2009} approximate the posterior to the non-linear system
over a GP using an MCMC sampling approach. Approaches to non-linear
likelihoods with GP priors, not limited to MOGPs, include the warped
GP model \citep{Lazaro2012} and chained GPs \citep{Saul2016} which
make use of variational approximations. Techniques from state space
modelling, including Taylor series linearisation and sigma points used
to approximate Gaussians in the extended and unscented Kalman filters
respectively have been applied to non-linear Gaussian processes, both
for single output and multitask learning \citep{Steinberg2014,
  Bonilla2016}.

% ####
% Possibly rewrite this with more focus on Hammerstein-Wiener models
% \/
An alternative perspective to the linear convolution process, in
particular latent force models, is to construct it as a
continuous-discrete state space model (SSM) driven by white noise
\citep{Hartikainen2012, Sarkka2013}. \citet{Hartikainen2012b} use this
approach for the general case of non-linear Wiener systems,
approximating the posterior with an unscented versions of the Kalman
filter and Rauch-Tung-Streibel smoother. The SSM approach benefits
from inference being performed in linear time, but relies on certain
constraints on the nature of the underlying covariance functions. In
particular, a kernel must have a rational power spectrum to be used in
exact form, which precludes the use of, for example, the exponentiated
quadratic kernel for exact Gaussian process regression without
introducing an additional approximation error
\citep{Sarkka2013}. \citet{Wills2013} also use a state space
representation to approximate Hammerstein-Wiener models, albeit with
sequential Monte Carlo and a maximum-likelihood approach.

\section{IMPLEMENTATION}\label{sec:implement}

\paragraph{Multi-output regression with NCMOGP} In this paper, we are
interested in the multi-output regression case. Therefore, we restrict
the likelihood models for each output to be Gaussian. In particular,
we assume that each observed output $y_d(t)$ follows $y_d(t) = \tilde{f}_d^{(C)}(t) + w_d(t)$,
where $w_d(t)$ is a white Gaussian noise process with covariance function $\sigma_d^2\delta_{t, t'}$. Other types of likelihoods are
possible as for example in \citet{MorenoMunoz:HetMOGP:2018}.

\paragraph{Kernel functions} In all the experiments, we use an exponentiated quadratic (EQ) form for the smoothing kernels
$G_d(t-\tau)$ and an EQ form for the kernel of the latent functions $u(t)$. With these forms, the kernel
$k_{d,d'}(t,t')$ also follows an EQ form. We use the expressions for $k_{d,d'}(t,t')$ obtained by
\citet{Alvarez:CompEfficient:jmlr:2011}.

\paragraph{High-dimensional inputs}
The resulting mean function $\ex[f_d^{(C)}(t)]$ and covariance function $\cov[f_d^{(C)}(t), f_{d'}^{(C)}(t')]$ assume that the input space is
one-dimensional. We can extend the approach to high-dimensional inputs, $\mathbf{x}\in \mathbb{R}^p$ by assuming that both the mean function and covariance function
factorise across the input dimension, and using the same expressions for the kernels for each factorised dimension.

\paragraph{Hyperparameter learning} We optimise the log-marginal likelihood for finding point estimates of the hyperparameters $\bm{\theta}$ of
the NCMOGP. Hyperparameters include the parameters for the smoothing kernels $G_d(\cdot)$, the kernel function $k(t,t')$ and the variances for the
white noise processes $w_d(t)$, $\sigma_d^2$. For simplicity in the notation, we assume that all the outputs are evaluated at the same set of inputs $\boldt = \{t_n\}_{n=1}^N$. Let $\boldy = [\boldy_1^{\top},\cdots, \boldy_D^{\top}]^{\top}$, with
$\boldy_d=[y_d(t_1),\cdot, y_D(t_N)]^{\top}$. The log-marginal likelihood $\log p(\boldy|\boldt)$ is then given as
\begin{align*}
   - \frac{ND}{2}\log(2\pi) -\frac{1}{2}  (\boldy-\bm{\mu}^{(C)})^{\top}(\boldK_{\boldf,\boldf}^{(C)}
    +\boldSigma)^{-1}  (\boldy-\bm{\mu}^{(C)}) -\frac{1}{2}\log\left| \boldK_{\boldf,\boldf}^{(C)} +
                              \boldSigma \right|,
\end{align*}
where $\bm{\mu}^{(C)}\in \mathbb{R}^{ND\times 1}$ has entries given by $\mu_d^{(C)}(t)$,
$\boldK_{\boldf,\boldf}^{(C)}\in \mathbb{R}^{ND\times ND}$ has entries computed using
$k_{d,d'}^{(C)}(t,t')$ and $\boldSigma$ is a diagonal matrix containing the variances of the noise
processes per output. We use a gradient-based optimization procedure to estimate the hyperparameters
that maximize the log-marginal likelihood. Computational complexity for this model grows as $\mathcal{O}(D^3N^3)$, related to the inversion of the matrix
$\boldK_{\boldf,\boldf}^{(C)}+\boldSigma$. This is the typical complexity in a full multi-output Gaussian process model \citep{Alvarez:CompEfficient:jmlr:2011}.

\paragraph{Predictive distribution} The predictive distribution is the same one used for the single-output case. Let $\boldt_*=[t_{n, *}]_{n=1}^{N_*}$ be the input test set. The
predictive distribution follows as $p(\boldy_*|\boldy) = \mathcal{N}(\boldy_*|\bm{\mu}_{\boldy_*|\boldy}, \boldK_{\boldy_*|\boldy})$, with $ \bm{\mu}_{\boldy_*|\boldy}
= \bm{\mu}^{(C)}_{*} + \boldK_{\boldf_*,\boldf}^{(C)}(\widetilde{\boldK}_{\boldf,\boldf}^{(C)})^{-1} \bm{\mu}^{(C)}$ and $\boldK_{\boldy_*|\boldy} = \boldK_{\boldf_*,\boldf_*}^{(C)} - \boldK_{\boldf_*,\boldf}^{(C)}(\widetilde{\boldK}_{\boldf,\boldf}^{(C)})^{-1}\boldK_{\boldf_*,\boldf}^{(C)\top}+\bm{\Sigma}_*$, where $\widetilde{\boldK}_{\boldf,\boldf}^{(C)} = \boldK_{\boldf,\boldf}^{(C)}+\bm{\Sigma}$. In these expressions, $\bm{\mu}^{(C)}_{*} $ has entries given
by $\mu^{(C)}(t_{n,*})$; $\boldK_{\boldf_*,\boldf_*}^{(C)} $ has entries given by $k_{d,d'}^{(C)}(t_{n,*}, t_{m,*})$; and  $\boldK_{\boldf_*,\boldf}^{(C)} $ has entries given by
$k_{d,d'}^{(C)}(t_{n,*}, t_{m})$.

% =======
% $\boldy_d=[y_d(t_1),\cdot, y_D(t_N)]^{\top}$. The marginal log-marginal likelihood is then given as
% \begin{align}
% \log p(\boldy|\boldt) =& - \frac{ND}{2}\log(2\pi)-\frac{1}{2}  \boldy^{\top}(\boldK_{\boldf,\boldf}^{(C)}+\boldSigma)^{-1} \boldy\notag\\
%                        &-\frac{1}{2}\log\left| \boldK_{\boldf^{(C)}, \boldf} + \boldSigma \right|,\label{eq:full:marginal}
% \end{align}
% where $\boldSigma$ is a diagonal matrix containing the variances of the noise processes per output, and
% $\boldK_{\boldf^{C},\boldf^{C}}\in \mathbb{R}^{ND\times ND}$ has entries computed using $k_{d,d'}^{(C)}(t,t'))$.
%
%
% As it is usual, we can use a gradient-based optimization procedure to estimate the hyperparameters that
% maximize the log-marginal likelihood leading to the infamous computational complexity of $\mathcal{O}(D^3N^3)$.

\section{EXPERIMENTAL RESULTS}

Experimental results are provided for the NCMOGP with homogeneous and separable kernels. In all the experiments that
follow, hyperparameter estimation is performed through maximization of the log-marginal likelihood as explained in section
\ref{sec:implement}. We use the normalised mean squared-error (NMSE) and the negative log-predictive density
(NLPD) to assess the performance.

\begin{figure*}[ht!]
\centering
\subfigure[Prediction for $y_1(t)$ with $C=1$]{ \label{fig:d:1:C:1}
\resizebox{0.32\textwidth}{!}{\includegraphics{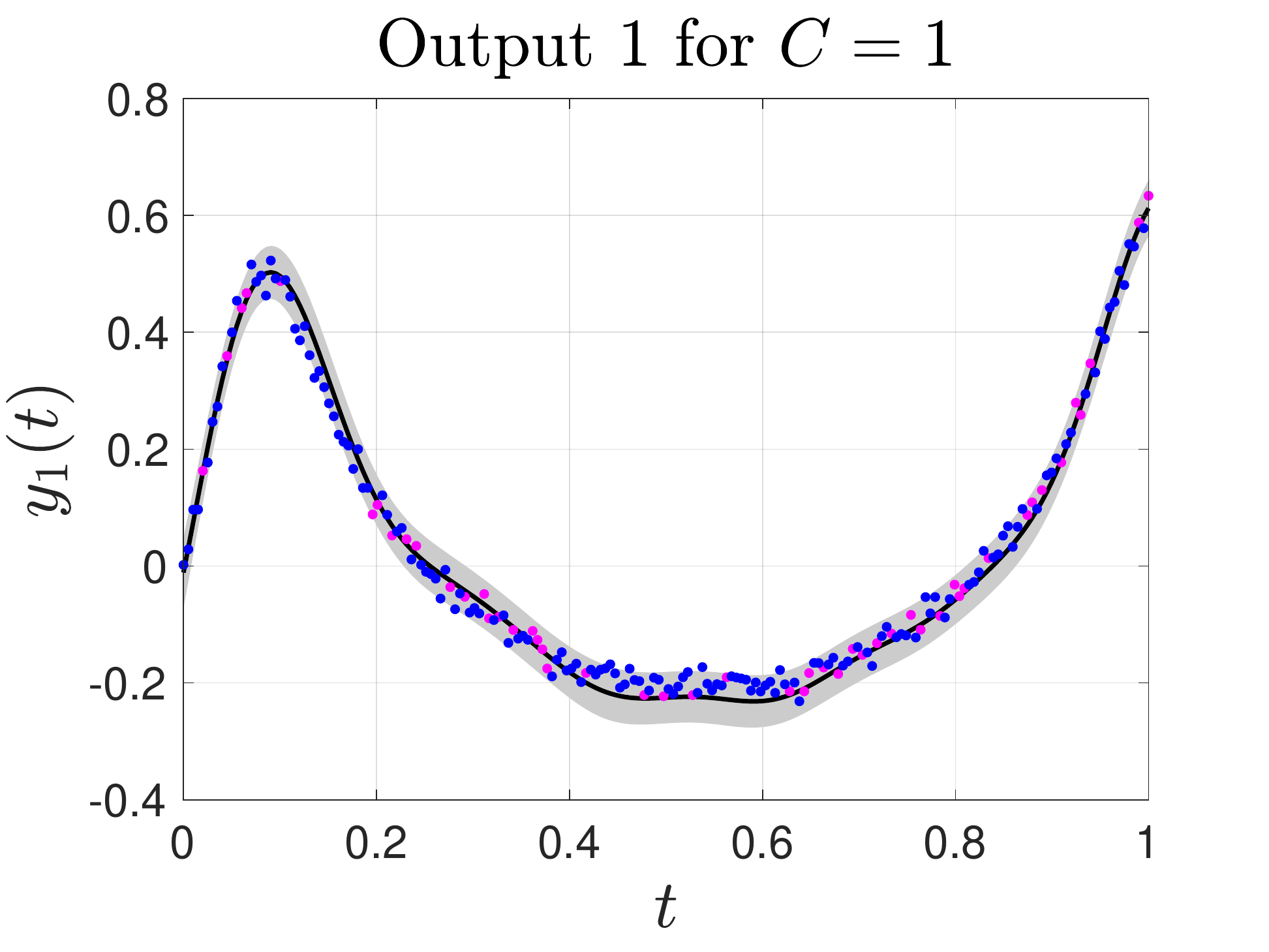}}}
\subfigure[Prediction for $y_2(t)$ with $C=1$]{ \label{fig:d:2:C:1}
\resizebox{0.32\textwidth}{!}{\includegraphics{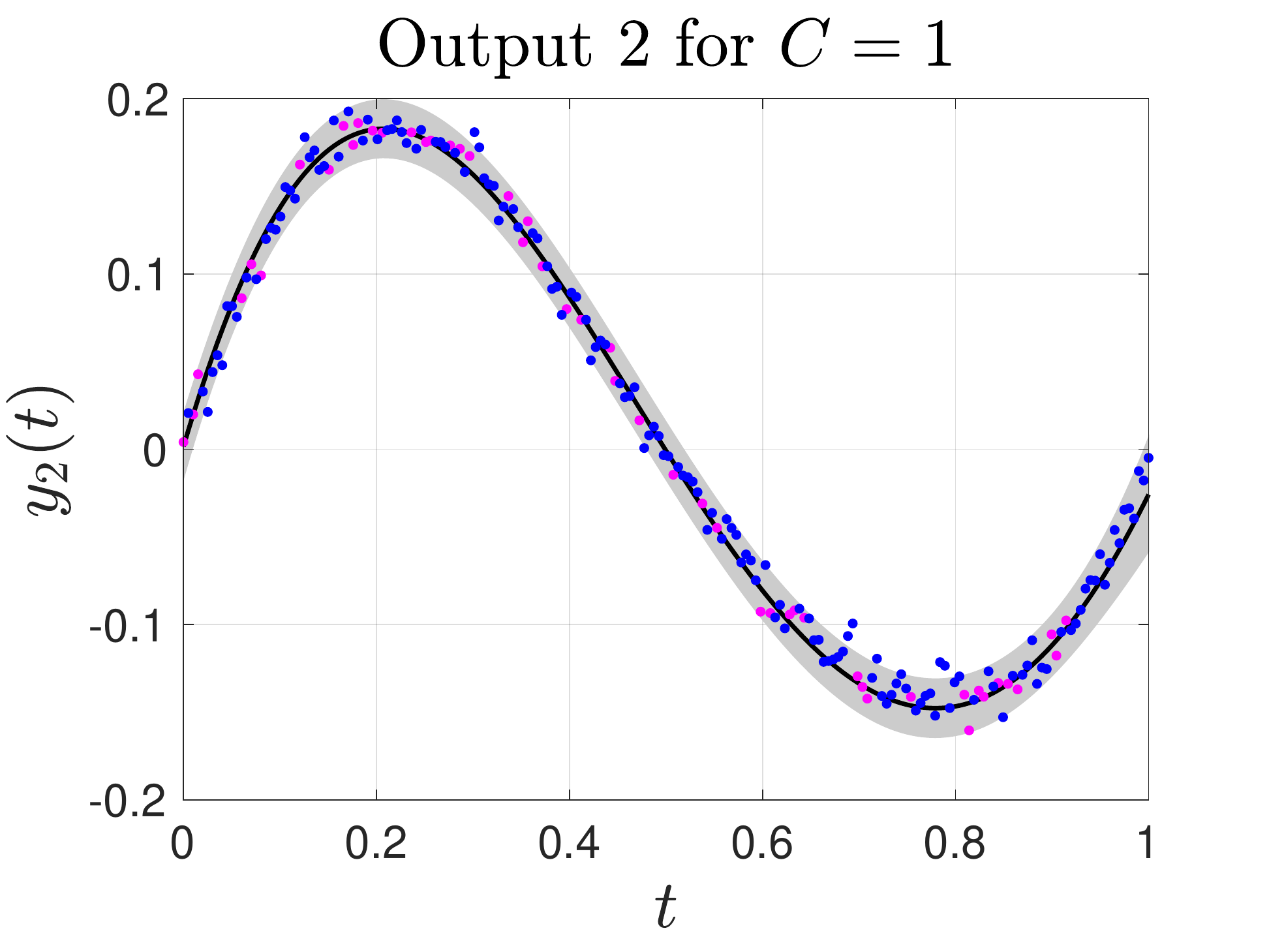}}}
\subfigure[Prediction for $y_3(t)$ with $C=1$]{\label{fig:d:3:C:1}
\resizebox{0.32\textwidth}{!}{\includegraphics{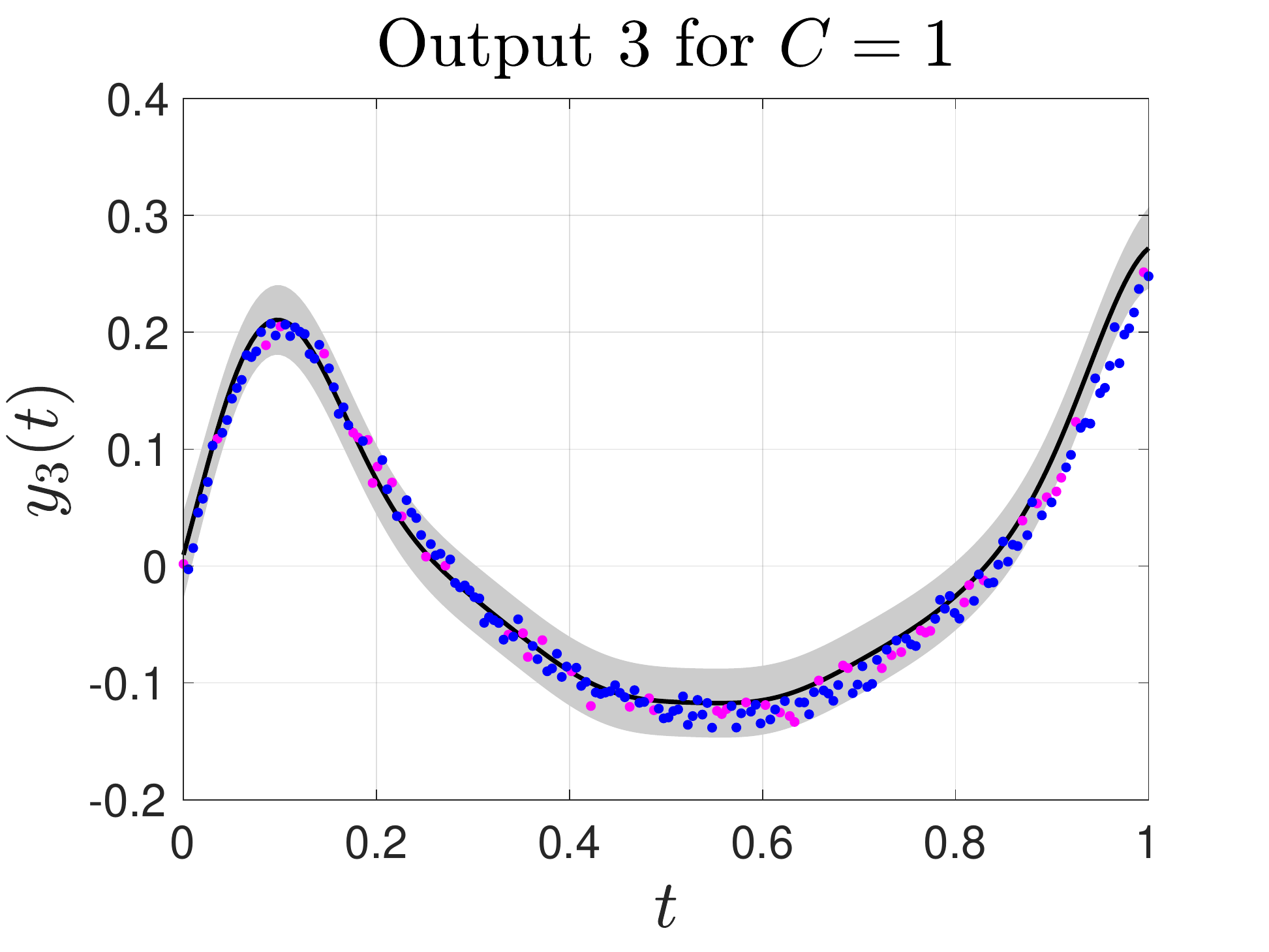}}}\\
\subfigure[Prediction for $y_1(t)$ with $C=3$]{\label{fig:d:1:C:3}
\resizebox{0.32\textwidth}{!}{\includegraphics{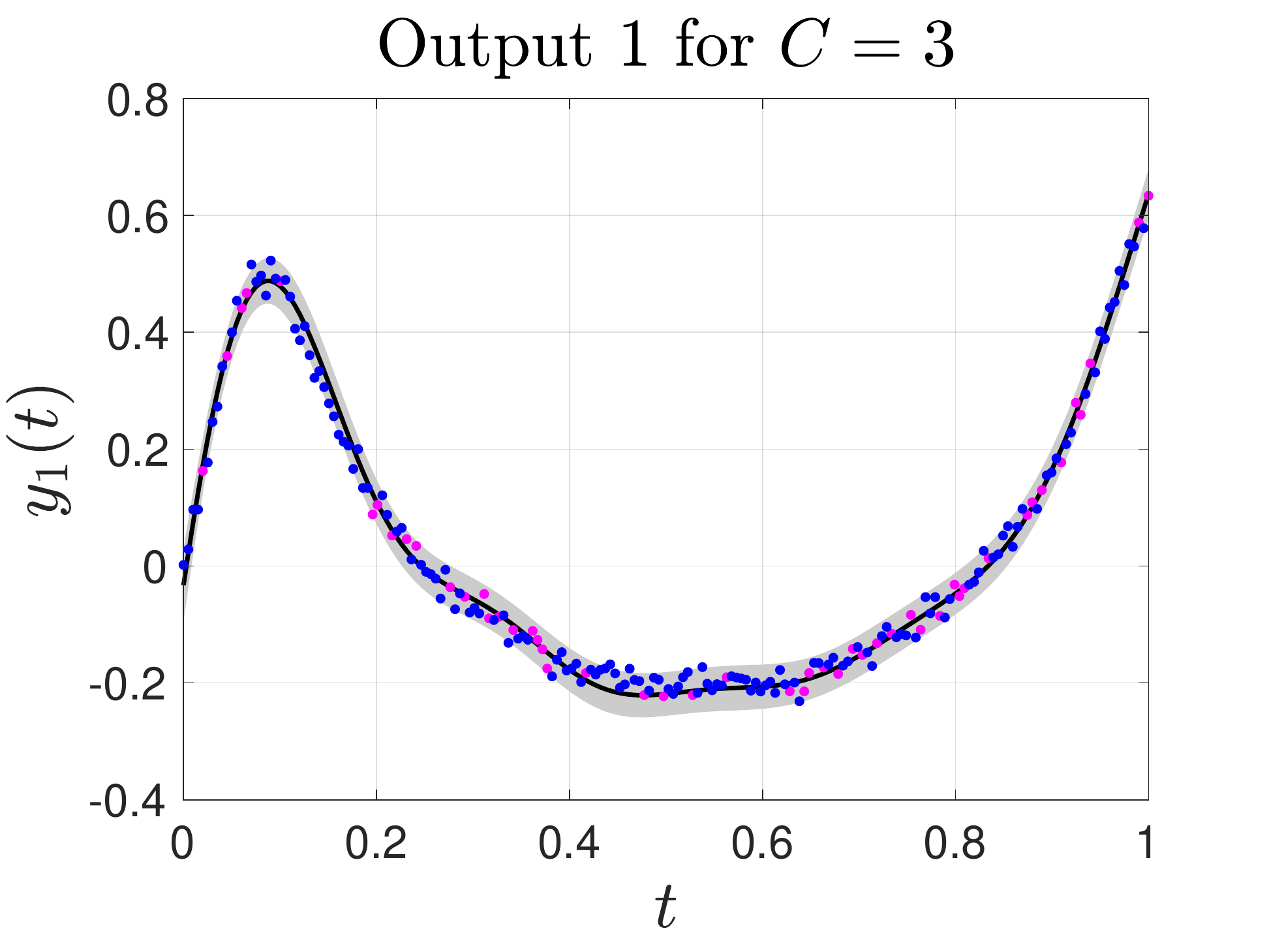}}}
\subfigure[Prediction for $y_2(t)$ with $C=3$]{\label{fig:d:2:C:3}
\resizebox{0.32\textwidth}{!}{\includegraphics{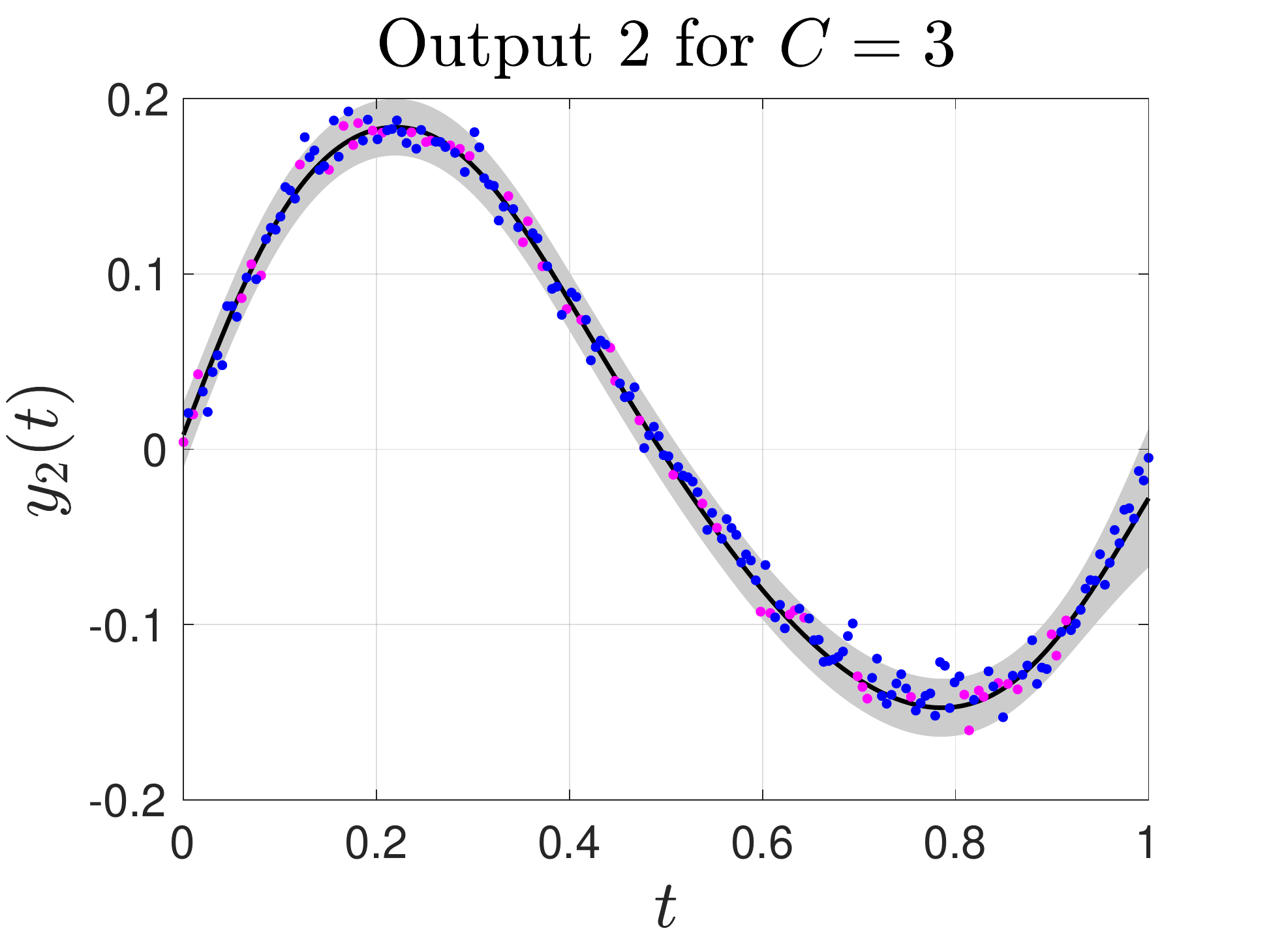}}}
\subfigure[Prediction for $y_2(t)$ with $C=3$]{\label{fig:d:3:C:3}
\resizebox{0.32\textwidth}{!}{\includegraphics{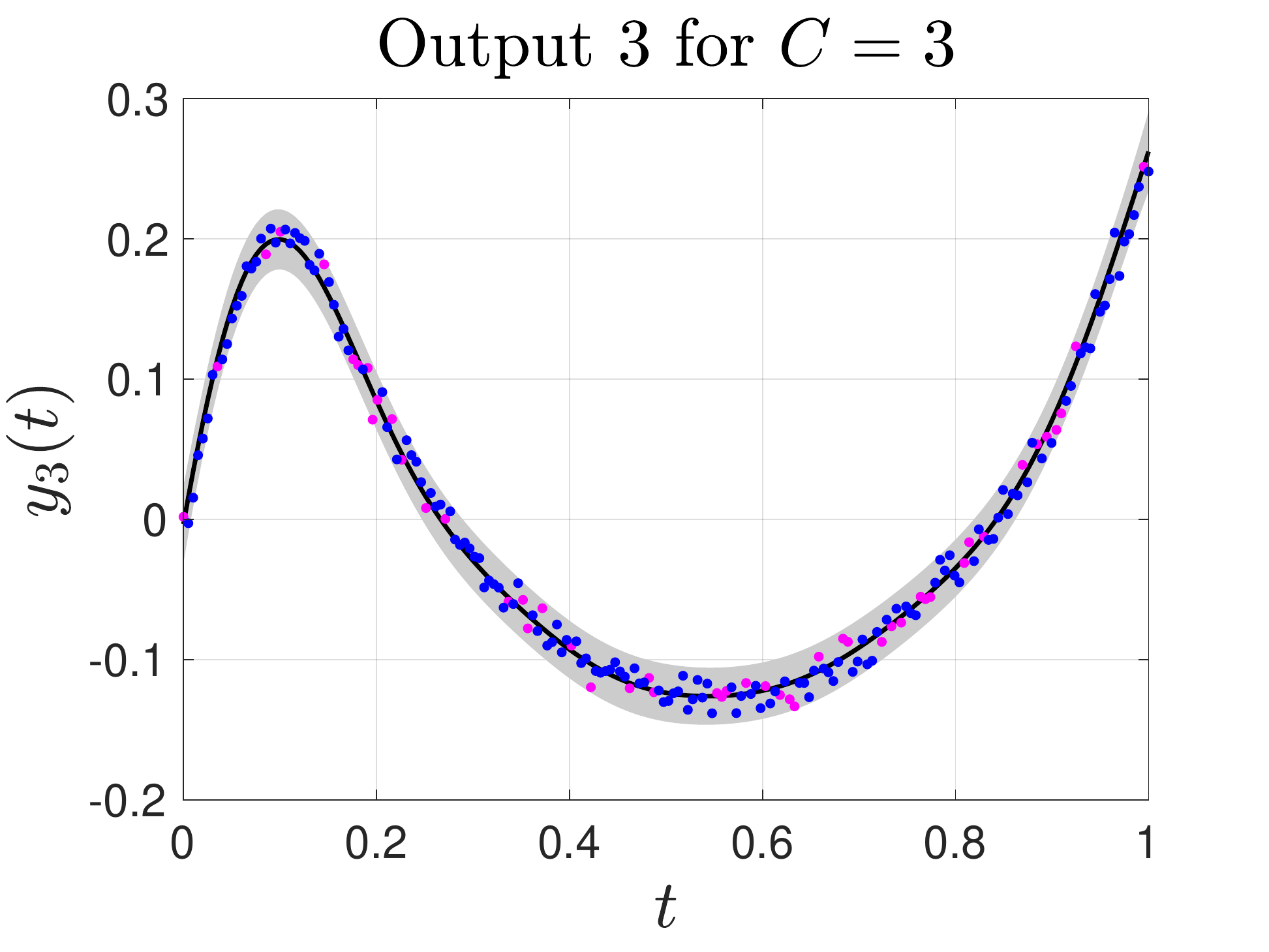}}}
\caption{Comparison of the predictive performance of CMOGP (top row) and NCMOGP with $C=3$ (bottom row)
  for the dataset synthetically generated. Each column is a different output. Training data is represented using
  red dots and Test data using blue dots. The black line in the mean over the predictive GP function, and the shaded region denotes two times the standard deviation.}
\label{fig:synthetic}
\end{figure*}

\subsection{Toy example}

We set a problem of $D=3$ outputs where the smoothing kernels are given as
$G_{d}(t-\tau)=S_d\exp(-P_d(t-\tau)^2)$, with parameters $S_1=5, S_2=1, S_3=2$, and $P_1=200, P_2=0.1$ and
$P_3=100$. The latent function follows as $u(t) =\sum_{k=1}^4\left(\frac{1}{k^2}\right)\cos(2k\pi t)$.  We then numerically solve the convolution integral $f_d(t) = \int_0^t G_d(t-\tau)u(\tau)d\tau$ for
$200$ datapoints in the input range $[0, 1]$. We compute $f_d^{(C=3)}(t) = \sum_{c=1}^3 f_d^c(t)$ and the
observed data is obtained by adding Gaussian noise with a variance of
$\sigma_d^2 = 0.005\times \operatorname{var}\left[f_d^{(C=3)}(t)\right]$. We randomly split the dataset
into a train set of $N=50$ per output and the rest of the datapoints are used for assessing the performance.

Figure \ref{fig:synthetic} shows qualitative results for predictions
made by the CMOGP and a NCMOGP with $C=3$. Although the outputs
exhibit a smooth behavior, there are subtle non-linear characteristics
that are not captured by the linear model. By zooming in on the
interval $t\in[0.4, 0.6]$ for output $y_1(t)$ (first column in the
Figure), we observe that the predictive mean of the linear model
($C=1$) does not completely follow the data compared to the true non-linear model
($C=3$). For output $y_3(t)$ (third column in the Figure), we observe
a similar behavior in the predictive mean of the linear model in the
intervals $t\in[0.4,0.8]$ and $t>0.9$ compared to the predictive mean
of the non-linear model.

Table \ref{tab:toy} shows the NMSE and the NLPD for different values of $C$ for twenty different partitions of the original dataset
into training and testing. We show the average of the metric plus-or-minus a standard deviation. The performance is similar for the non-linear models with $C\ge 3$, although the model for $C=3$ shows the
lowest standard deviation.

\begin{table}[ht!]
\caption{Results for the Toy example.}
\label{tab:toy}
\centering
\begin{tabular}{|c|c|c|}\hline
$C$ & NMSE & NLPD  \\\hline
  1   &  $0.0142\pm 0.0022$ & $-2.6868\pm 0.0794$ \\
  2   &  $0.0075 \pm 0.0012$ & $-2.9641\pm 0.0665$\\
  3   &  $\mathbf{0.0071 \pm 0.0009}$ & $\mathbf{-2.9780\pm 0.0513}$\\
  4   &  $ 0.0072\pm 0.0010$ & $-2.9546\pm 0.0960$\\
  5   &  $0.0071\pm 0.0011$&  $-2.9428\pm 0.0956$\\\hline
\end{tabular}
\end{table}

\subsection{Weather data}
We use the air temperature dataset considered previously by \citet{Nguyen:2014} and used in other multi-output GP papers. The dataset contains air temperature measurements at four spatial locations on the south coast of
England: Bramble Bank, Southhampton Dockhead, Chichester Harbour and Chichester Bar, usually refer to by the names
Bramblemet, Sotonmet, Cambermet and Chimet, respectively. The measurements correspond to July 10 to July 15, 2013. The
prediction problem as introduced in \citet{Nguyen:2014} corresponds to predicting consecutive missing data in the outputs
Cambermet and Chimet, $173$ and $201$ observations, respectively, using observed data from all the stations: $1425$
observations for Bramblemet, $1268$ observations for Cambermet, $1235$ for Chimet and $1097$ for Sotonmet.

% \subsection{MOCAP data}
% We use two motion examples from the CMU MOCAP database\footnote{The CMU MOCAP dataset is available at
%   \url{http://mocap.cs.cmu.edu/}.}  to illustrate the performance of the models. The motion examples are
% from subject 02: a walking (trial 01) and a running (trial 03). For both movements we use $D=10$ outputs\footnote{The outputs were\texttt{'rhumerus_X','rhumerus_Y','rradius','lhumerus_X','lhumerus_Y',...
%  'lradius_X','rfemur_X','rfoot_X','lfemur_X','lfoot_X'}}
The missing data was artificially removed and we have access to the
ground-truth measurements. We compare the non-linear convolution
approach to the Intrinsic Coregionalization Model (ICM)
\citep{Goovaerts:book97} and Dependent Gaussian processes (DGP)
\citep{Boyle:dependent04}. Both models can be seen as particular cases
of the non-linear model in Eq. \eqref{eq:finite:ncmogp}. The ICM can
be recovered from Eq. \eqref{eq:finite:ncmogp} by making $C=1$ and
$G_{d,1}(\cdot)=a_{d,1}\delta(\cdot)$, where $a_d$ is a scalar that
needs to be estimated and $\delta(\cdot)$ is the Dirac delta
function.\footnote{Notice that we use a rank one ICM since, similarly
  to the non-linear model, it only involves one latent function
  $u(\cdot)$ \citep{alvarez2012kernels}} DGP is also recovered from
\eqref{eq:finite:ncmogp} assuming $C=1$ and a white-Gaussian process
noise model for $u(\cdot)$.

Table \ref{tab:weather} reports the mean predictive performance for the missing observations for five random
initialisations of the models. CMOGP refers to the NCMOGP with $C=1$. The number at the end of NCMOGP indicates the
order $C$ of the non-linear model. Results are shown per output with missing data. Notice the reduction in the
NMSE for Cambermet for all the non-linear models compared to the linear models (CMOGP and DGP). Here ICM performs better than any of the convolution approaches. For Chimet, there is also a reduction in the NMSE, but only
for $C=3$ (NCMOGP3) and $C=5$ (NCMOGP5). In terms of the NLPD, the non-linear model with $C=5$ (NCMOGP5) has a competitive performance when compared to the
linear model (CMOGP). The averaged NSMEs for the two outputs are:
$0.6989$ ($C=1$), $0.5717$ ($C=2$), $0.4309$ ($C=3$), $0.5456$
($C=4$), $0.4753$ ($C=5$), $0.4425$ (ICM) and $0.6431$ (DGP). The averaged NLPDs for the two ouputs are: $2.0613$ ($C=1$), $3.2479$ ($C=2$), $2.2550$
($C=3$), $2.5460$ ($C=4$), $1.9600$ ($C=5$), $2.6848$ (ICM) and $1.9925$ (DGP). NCMOGP5 offers the best competitive performance both in terms of the NMSE and the NLPD compared to CMOGP, ICM and DGP.

\begin{table}[ht!]
\caption{Results for the Weather dataset.}
\label{tab:weather}
\begin{center}
  \begin{tabular}{|c|c|c|c|c|}\hline
 \multirow{2}{*}{Model} & \multicolumn{2}{c|}{Cambermet} & \multicolumn{2}{c|}{Chimet} \\\cline{2-5}
                               & NMSE & NLPD & NMSE & NLPD   \\\hline
  CMOGP    &  $0.6004$ & $1.9830$    & $0.7974$ & $2.1397$\\
  NCMOGP2   &  $0.3080$ & $2.1604$ & $0.8354$ & $4.3355$\\
  NCMOGP3   &  $0.4366$ & $2.3333$& $\mathbf{0.4252}$ & $2.1766$\\
  NCMOGP4   &  $0.2676$ & $2.1075$& $0.8235$ & $2.9846$\\
  NCMOGP5   &  $0.4499$ & $2.0919$& $0.5007$& $\mathbf{1.8280}$\\
     ICM         &  $\mathbf{0.1048}$ & $3.2211$& $0.7801$& $2.1485$\\
     DGP        &  $0.5513$ & $\mathbf{1.9463}$& $0.7349$& $2.0388$\\\hline
\end{tabular}
\end{center}
\end{table}

\subsection{A high-dimensional input example}

The NCMOGP can also be applied for datasets with an input dimension greater than one. We use a subset of the SARCOS dataset\footnote{Available at  \url{http://www.gaussianprocess.org/gpml/data/}} for illustration purposes. The prediction problem corresponds to map from positions, velocities and accelerations to the joint torques in seven degrees-of-freedom SARCOS anthropomorphic
robot arm. The datasets contains $D=7$
outputs and the dimension of the input space is $p=21$, corresponding to seven positions, seven velocities, and seven accelerations.
The kernels that we use follow the idea described in Section~\ref{sec:implement} for higher-dimensional inputs.

Our setup is as follows:
from the file \texttt{sarcos\_inv.mat} we randomly select $N=500$ observations for each output for training and from the file
\texttt{sarcos\_inv\_test.mat}, we randomly select another $500$ observations for testing. We repeat the experiment ten times for
different training and testing sets taken from the same two files.

\begin{table}[h]
\caption{Results for a subset of the SARCOS dataset.}
\label{tab:sarcos}
\begin{center}
\begin{tabular}{|c|c|c|}\hline
Model & NMSE & NLPD  \\\hline
  CMOGP   &  $0.0497\pm0.0252$ & $1.4292\pm1.1080$\\
  NCMOGP2   &  $0.0478\pm0.0238$ & $\mathbf{1.4067\pm1.1032}$\\
  NCMOGP3   &  $0.0571\pm0.0377$ & $1.4164\pm1.1401$\\
  NCMOGP4   &  $0.0720\pm0.0855$ & $1.4294\pm1.0787$\\
  NCMOGP5   &  $0.0830\pm0.0809$ & $1.4674\pm1.1449$\\
  ICM   &  $0.0513\pm0.0246$ & $1.4208\pm1.1286$\\
  DGP   &  $\mathbf{0.0477\pm0.0237}$ & $1.4221\pm1.1088$\\\hline
\end{tabular}
\end{center}
\end{table}

Table \ref{tab:sarcos} shows the averaged NMSE and averaged NLPD for
the ten repetitions plus-or-minus a standard deviation. The non-linear
model NCMOGP2 yields better averaged performance than CMOGP and ICM,
in terms of NMSE and a very similar performance to DGP. In terms of
NLPD, NCMOGP2 outperforms all the other models, indicating a better
performance in terms of the predicted variance.  Something to point
out is that when looking at the predictive performance for the
different NCMOGP, we noticed that each output is usually better
predicted by different values of $C$. For example, in terms of NLPD,
the best order to predict outputs $d=1,2,3,7$ would be a non-linear
model with $C=4$.  The best model for predicting outputs $d=4, 6$
would be $C=3$, and the best model for predicting output $d=5$ would
be $C=5$.

\section{CONCLUSIONS AND FUTURE WORK}

We have introduced a non-linear extension of the
process convolution formalism to build multi-output Gaussian
processes. We derived a novel mean function and covariance function from the non-linear
operations introduced by the transformations in a Volterra series and
showed experimental results that corroborate that these non-linear
models have indeed an added benefit in real-world datasets.

We envision several paths for future work. The most pressing one is
extending the framework to make it suitable for larger datasets. We
can use similar ideas to the ones presented in
\citet{MorenoMunoz:HetMOGP:2018} to establish a stochastic
variational lower bound for the model introduced in this
paper. Exploring the non-linear models for the case of latent force
models is also an interesting venue. In this paper we use an smoothing
kernel with an EQ form, but it is also possible to use smoothing
kernels that correspond to Green's function of dynamical
systems. Automatically learning the smoothing kernel from data is also
an alternative as for example in \citet{Guarnizo:2017}.

An observation from both the Weather and SARCOS experiments, one that could have been
expected, is that the predictive performances for the outputs are not all equally good for the
same value of $C$. A potential extension of our model would be to
allow the automatic learning of the order $C$ per output dimension,
say $C_d$. Additionally, the kernels we considered were derived assuming that the Volterra
kernels were separable and homogeneous. Relaxing both assumptions is
yet another path for future research.

\bibliographystyle{plainnat}
\bibliography{nonlinconv}

\clearpage
\appendix
{\huge\bf Supplementary Material}

\section{LOG LIKELIHOOD}
For a given vector of inputs, $\mathbf{t}=[t_n]^N_{n=1}$, the log-marginal likelihood is given as
\begin{align*}
    \log p(\mathbf{y}\,|\,\mathbf{t}) = -\frac{ND}{2}\log(2\pi) -\frac{1}{2}\log|\boldK_{\boldf,\boldf}^{(C)} + \bm{\Sigma} - \frac{1}{2}(\mathbf{y}-\bm{\mu}^{(C)})^\top(\boldK_{\boldf,\boldf}^{(C)} + \bm{\Sigma})^{-1}(\mathbf{y}-\bm{\mu}^{(C)}),
\end{align*}
where $\mathbf{y} = [\mathbf{y}_1^\top\,\ldots\,\mathbf{y}_D^\top]^\top$ with $\mathbf{y}_d = [y_d(t_1)\,\ldots\,y_d(t_N)]^\top$; $\bm{\mu}^{(C)}$ and $\boldK_{\boldf,\boldf}^{(C)}$ have entries given by $\mu_d^{(C)}(t)$ and $k_{d,d'}^{(C)}(t,t')$ respectively; and $\bm{\Sigma}$ is the diagonal matrices containing the variances of the noise processes per output. We can find derivatives to this much in the same way as the single output \citep{Rasmussen2006}.

\section{DERIVATIVES W.R.T. HYPERPARAMETERS}
We provide here the derviatives of the mean and covariance functions for the output process with respect to hyperparameters for the purpose of optimisation using a gradient-based method.

\subsubsection*{Mean Function for Output Process}
The mean function for output process is defined
\begin{align*}
    \mu_{f_d}(t) = \sum^C_{c=1}\frac{c!}{2^{c/2}\left(\frac{c}{2}\right)!}\left(k_{d,d}(t,t)\right)^{c/2}
\end{align*}
and its derivative with respect to the kernel hyperparameters $\theta^d$ is
\begin{align*}
    \frac{\partial\mu_{f_d}(t)}{\partial\theta_d} = \sum^C_{c=1}\frac{c!}{2^{(c/2)}\left(\frac{c}{2}\right)!}\left(\frac{c}{2}\right)\left(k_{d,d}(t,t)\right)^{(c/2)-1}\frac{\partial k_{d,d}(t,t)}{\partial \theta^d}.
\end{align*}

\subsubsection*{Covariance Function for Output Process}
We define the covariance function of the output process as
{\small%
\begin{align*}
    k_{y_d,y_{d'}}(t,t') =& \sum^C_{c=1}\sum^C_{c'=1}\sum_{\mathbf{L}\in T_\mathbf{a}}A_{c,c',\mathbf{L}}\bigg((k_{d,d}(t,t))^{l_{11}}(k_{d,d'}(t,t'))^{l_{12}}(k_{d',d'}(t',t'))^{l_{22}}\bigg)\\&\quad-\Bigg(\sum^C_{c=1}\frac{c!}{2^{c/2}\left(\frac{c}{2}\right)!}(k_{d,d}(t,t))^{c/2}\Bigg) \Bigg(\sum^C_{c'=1}\frac{c'!}{2^{c'/2}\left(\frac{c'}{2}\right)!}(k_{d',d'}(t',t'))^{c'/2}\Bigg),
\end{align*}}%
where
\begin{align*}
    A_{c,c',\mathbf{L}} = \frac{c!c'!}{2^{l_{11}+l_{22}}l_{11}!l_{12}!l_{22}!}.
\end{align*}

The derivative with respect to $\theta_d$ is therefore
{\small%
\begin{align*}
    \frac{\partial k_{y_d,y_{d'}}(t,t')}{\partial \theta^d} =&
    \sum^C_{c=1}\sum^C_{c=1}\sum_{\mathbf{L}\in T_{\mathbf{a}}} A_{c,c',\mathbf{L}}\Bigg(l_{11}(k_{d,d}(t,t))^{l_{11}-1}(k_{d,d'}(t,'t))^{l_{12}}(k_{d',d'}(t',t'))^{l_{22}}\frac{\partial k_{d,d}(t,t)}{\partial \theta^d} \Bigg)\\&\quad+\Bigg(l_{12}(k_{d,d}(t,t))^{l_{11}} (k_{d,d'}(t,'t))^{l_{12}-1}(k_{d',d'}(t',t'))^{l_{22}}\frac{\partial k_{d,d'}(t,t')}{\partial \theta^d} \Bigg)\\&\quad- \Bigg(\sum^C_{c=1}\frac{c!}{2^{c/2}\left(\frac{c}{2}\right)!}\left(\frac{c}{2}\right)(k_{d,d}(t,t'))^{(c/2)-1}\frac{\partial k_{d,d}(t,t)}{\partial \theta^d}\Bigg)\Bigg(\sum^C_{c'=1}\frac{c'!}{2^{c'/2}\left(\frac{c'}{2}\right)!}(k_{d',d'}(t',t'))^{c'/2}\Bigg).
\end{align*}}%

\end{document}